\documentclass[letterpaper]{article}
\usepackage{iccc}

\usepackage{times}
\usepackage{latexsym}
\usepackage{graphicx}
\usepackage{url}
\usepackage{amsmath}
\usepackage{amssymb}
\usepackage{xcolor}
\usepackage{subcaption}

\usepackage{times}
\usepackage{helvet}
\usepackage{courier}
\title{Toward Automated Quest Generation in Text-Adventure Games \\
\author{Prithviraj Ammanabrolu \:\:\:\: William Broniec  \:\:\:\: Alex Mueller \:\:\:\: Jeremy Paul \:\:\:\: Mark O. Riedl\\
School of Interactive Computing\\
Georgia Institute of Technology\\
Atlanta, GA 30332  USA\\
\texttt{raj.ammanabrolu@gatech.edu}\\
}
}
\newcommand{\citet}[1]{\citeauthor{#1}~\shortcite{#1}}

\begin{document} 
\maketitle

\begin{abstract}
\begin{quote}
Interactive fictions, or text-adventures, are games in which a player interacts with a world entirely through textual descriptions and text actions.
These games are typically structured as puzzles or quests wherein the player must execute certain actions in a certain order to succeed.
In this paper, we consider the problem of procedurally generating a quest, defined as a series of actions required to progress towards a goal, in a text-adventure game.
Quest generation in text environments is challenging because they must be semantically coherent. 
We present and evaluate two quest generation techniques: 
(1)~a Markov model, and
(2)~a neural generative model.
We specifically look at generating quests about cooking and train our models on recipe data.
We evaluate our techniques with human participant studies looking at perceived creativity and coherence.
\end{quote}

\end{abstract}

\section{Introduction}


Natural language can be used to express creativity in the form of narrative.
Prior research has shown that narrative is used in everything from environmental understanding \cite{Bruner1991-BRUTNC} to developing language \cite{Johnston}.
Given this wide ranging impact, using narrative in language to help us understand human perceptions of creativity and what it takes to replicate this through computational models is natural.
Text-adventure games, or interactive fiction, in which a player interacts with a world entirely through text, provide us with a platform on which to explore these ideas on creativity in language.
These games are usually structured as puzzles or quests in which a player must complete a sequence of actions in order to succeed.
Text games allow us to factorize the problem of creative language generation and focus on developing more fine-grained, data-driven models.

Automated generation of text-adventure games can broadly be split into two considerations: 
(1)~the structure of the world, including the layout of rooms, textual description of rooms, objects, and non-player characters; 
and
(2)~the quest, consisting of the partial ordering of activities that the player must engage in to make progress toward the end of the game.
In this work, we focus on methods of automatically generating such a quest and how it can be used to better understand narrative intelligence, specifically looking at perceived creativity and coherence.
Quest generation requires narrative intelligence as a quest must maintain coherence throughout and progress towards a goal.
Maintaining quest coherence also means following the constraints of the given game world.
The quest has to fit within the confines of the world in terms of both genre and given affordances---e.g. using magic in a fantasy world.
This is further complicated in the case of a text-adventure as a consequence of all interactions being in natural language---the potential output space is combinatorial in size.
Because the player ``sees'' and ``acts'' entirely through text, any quest generation system must also take into account the lack of visual information and generate sufficiently descriptive text accordingly.

There are multiple variables that could potentially affect a player's perception of creativity in a text-adventure game such as the vocabulary used, the structure of the world, stylistic variations in writing, etc.
To conduct controlled studies,
we use the TextWorld framework \cite{cote2018textworld} which lets us generate text-adventure game {\em worlds} based on a grammar, allowing us to focus on novel quest generation algorithms.
It lets us fix variables concerned with game world and logic generation and focus only on the generation of quests within this world.
We use this framework's ``home'' theme---providing us with a textual simulation of a house---and restrict the types of quests that can be generated to those involving the completion of a cooking recipe.
We then attempt to learn how to generate a quest to complete a recipe---as well as how to create the recipe itself---using a large scale knowledge base of recipes. 
In these quests, players are provided with a list of ingredients and their locations, and they have to navigate the environment to find and prepare those ingredients to complete the given recipe. 
For example, given a recipe to make peanut butter cookies the quest would first tell the player to find eggs, peanut butter, flour, and baking soda.
The player would then have to figure out that the first ingredient is in the fridge while the others are in the pantry and prepare each item accordingly.
Generating this sort of quest requires knowledge of the ingredients, how they fit together, and how those ingredients interact with the environment.

The contribution of this work is thus twofold.
We first detail a framework, and variations thereof, that can learn to generate creative quests in a text-adventure game.
This framework includes two quest generation models using Markov chains as well as a neural language model.
It also uses a semantically grounded knowledge graph to improve overall quest coherence.
Our second contribution provides human subject evaluations that give us insight into how each variation of this framework affects human perception of creativity and coherence in such games.  

\section{Related Work}

Although there has been much work recently on text-adventure gameplay \cite{bordes2010towards,He2015,Narasimhan2015,fulda-ijcai2017,mtdweston,Haroush2018,cote2018textworld,tao18,ammanabrolu,ammanabrolutransfer,jericho,hausknecht,ammanabrolu2020graph}, these works focus on creating agents that can play a given game as opposed to being able to automatically generate content for them.

%

Outside of this, there has been some work on learning to create content in the context of interactive narrative.
These systems mainly work to overcome a significant bottleneck in the form of the human authoring required to create such works.
\citet{Permar2013} present a method of generating cognitive scripts required for freeform activities in the form of pretend play.
Specifically, they use interactive narrative---a form of pretend play that requires a high level of improvisation and creativity and uses cognitive scripts acquired from multiple experience sources.
They take existing cognitive scripts and blend them in the vein of more traditional conceptual blending \cite{Veale,Zook11} to create new blended scripts.
Closely related is \citet{Magerko} who present a Co-Creative Cognitive Architecture (CoCoA), detailing the set of components that support the design of co-creative agents in the context of interactive narrative.
These methods all follow singular cognitive models that do not learn to generate content automatically.

\citet{Li2012} present Scheherazade, a system which learns a plot graph based on stories written by crowd sourcing the task of writing short stories through Amazon Mechanical Turk.
This plot graph contains details relevant for the coherence of the story and includes: plot events, temporal precedence, and mutual exclusion relations.
The generated narrative contains events that can be executed from this plot graph by both players and non-player characters.
\citet{guzdial2015crowdsourcing} introduce Scheherazade-IF, a system that learns to generate choose-your-own-adventure style interactive fictions in which the player chooses from prescribed options.
More recently, \citet{Martin2017a} introduce a pipeline systems for improvisational storytelling agents capable of collaboratively creating stories.
These agents first focus on creating a plot for the story and then expand that plot into natural language sentences.

\citet{Giannatos2011} use genetic algorithms to create new story plot points for an existing game of interactive fiction using an encoding known as a precedence-constraint graph.
This graph gives the system information regarding the ordering of events that must happen in the game in order to advance.
They demonstrate the workings of their system by generating additional content for the popular interactive fiction game {\em Anchorhead}, and show that this can be integrated into the original game.
This work, however, is offline and relies on existing interactive fiction games and having knowledge of the precedence-constraint graph for this existing game.

The Game Forge system \cite{hartsook} also uses genetic algorithms to generate a game world and plot line for related type of game, a computer role playing game (CRPG).
This work focuses on generating layouts and plot structures to create novel game worlds through
with a fitness function based on a transition graph that encodes pre-built game requirements.
\citet{tamari-etal-2019-playing} focus on extracting action graphs for sequential decision making problems such as material science experiments and turn them into text-adventure games.
Although these works use graph structures in order to constrain the generation of the game, we use these graph structures only to maintain coherence and focus on content creation.

Although there are works that attempt to automatically evaluate the creativity of the output of a generative process by computationally modeling potential human responses --- such as with story telling \cite{Purdy}, etc. --- we choose to rely on a human subject study based on the definition of creativity as presented in \citet{boden2007creativity}.
Specifically we focus on the concepts of novelty and value, despite collecting data for other defined metrics as well.
We use the definition of novelty stemming from the idea of p-creativity, i.e. a concept that is entirely new to a single agent -- in this case a subject in our evaluation study.
Value, as a component of computational creativity, however, is not defined concretely in Boden's work for a general domain.
Our definition of value in the context of text-adventure games relies on accomplishment or achievement.

\citet{ammanabrolu20world} approach the problem of world generation in interactive fiction by turning linear stories into interactive worlds.
They first extract a knowledge graph of the world from the story---containing lcoations, characters, and objects---and use that to generate the full game.
\citet{fan2019generating} leverage a crowdsourced dataset of fantasy text-adventure dialogues~\cite{Urbanek2019LearningTS} to learn to generate interactive fiction worlds on the basis of of locations, characters, and objects.
These works all focus on the problem of world generation in text-adventure games and do not contain objectives or quests---these systems are thus complimentary to ours.

\begin{figure}
    \centerline{\includegraphics[width=0.65\linewidth]{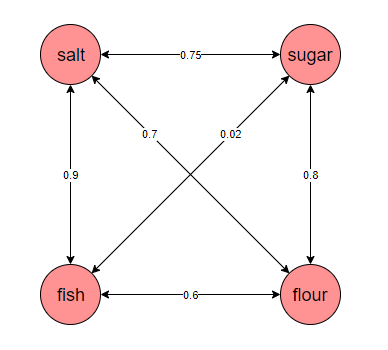}}
    \caption{Example of ingredient connections.}
    \label{ingredidentExample}
\end{figure}

\section{Content Generation}

In this section, we present Markov chains and neural language model based models to generate content, i.e. recipes, for our quests. 
Content generation for a quest in a text-adventure game, in this case a recipe, can be thought of as being equivalent to generating a sequence of events in which prior elements affect the probability of subsequent events.
Markov chains present a simplified and well studied method to generate such content.
Neural language models, designed to predict an element of a sequence conditioned on a given number of prior elements, let us generate sequences of events with more prior context---i.e. in the absence of the Markov assumption.

\subsection{Markov Chains}
Our first quest generation model is based on the use of Markov chains.
This generation process consists of two steps.
We first learn a weighted {\em ingredient graph}, a Markov chain, from a large scale knowledge base of recipes and then probabilistically walk along this graph to generate the instructions for the recipe.

\subsubsection{Ingredient Graph}
\label{sec:ingr}

\begin{figure}
    \centerline{\includegraphics[width=\linewidth]{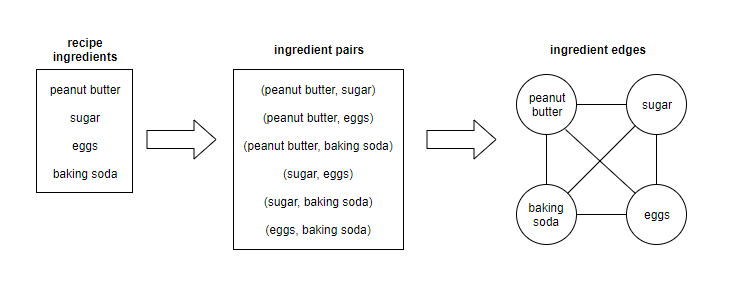}}
    \caption{Ingredient extraction process.}
    \label{ingredientExtraction}
\end{figure}

Generating the recipe requires domain knowledge.
For example, creating a recipe for peanut butter cookies requires an understanding that an ingredient like peanut butter fits well with eggs, flour, and sugar while something like fish does not. 
We represent this knowledge with an undirected graph of ingredients. 
Our ingredient graph is based off of recipes scraped from \url{allrecipes.com}.\footnote{\url{https://github.com/kbrohkahn/recipe-parser}}
The raw, uncleaned dataset included over 20,000 recipes with over 4000 unique ingredients.
A list of ingredients was extracted from each recipe, and each of these lists was converted into a set of ingredient pairs (Fig.~\ref{ingredientExtraction}). 
In total, there were 118,116 unique ingredient pairings, and 73,088 of those pairings (62\%) only occurred once.
We reduced the number of distinct ingredients from 4460 to 1703 by merging items with the same base ingredient and by replacing name-brand items with a generic equivalent.

Each of the nodes within the graph represents a possible ingredient, and weighted connections between these nodes represent how well the ingredients go together.
The weight of each edge is the total number of occurrences of that ingredient pair within the recipe corpus. 
The edge connecting eggs and white sugar would have a weight of 3774 while the edge between hot milk and orange juice would have a weight of 1. 
Ingredient pairings that do not occur within the recipe corpus did not have an edge within this network, and this helped prevent our model from generating completely incoherent recipe pairings (e.g. hot sauce and baby food).
Take the graph in Fig.~\ref{ingredidentExample} as an example.
In this complete graph, all of the ingredients go well with each other except for fish and sugar, which is indicated by the low weight connection between them. 
The weak connection between sugar and fish suggest that they would likely not go well together in a recipe.

\subsubsection{Instruction Generation}

With the ingredient graph created, we begin the process of instruction generation based on sub-graph mining and prior generative methods based on probabilistic graph walks \cite{fleishman1978method}. 
We start by selecting an initial random ingredient `x' weighted by its distribution in the graph.

\begin{align}
p(x_{1}) &= \frac{\sum_{i=1}^{k}{w(v_{i}, x_{1})}}{\sum_{i=1}^{k}{\sum_{j=1}^{k}{w(v_{i}, v_{j})}}}
\end{align}

We probabilistically select one of its neighbors based on the conditional frequency of the pair. 
Each iteration further computes conditional probabilities and selects them.
We exclude all ingredients in which any bag of words token is contained by any other, ensuring that a variety of different ingredients are selected.
\begin{align}
\alpha &= \begin{cases}0 & B_{x_{i}} \subseteq B_{x_{n+1}}  \vee B_{x_{n+1}} \subseteq B_{x_{i}}\\1 & else\end{cases}
\label{eq:bagofwords}
\end{align}
In Eq.~\ref{eq:bagofwords}, B\textsubscript{x\textsubscript{i}} refers to the 1-gram bag of words model.

However, just computing complete conditional probabilities would remove the chance for entirely new combinations to emerge. 
Therefore, we calculate just the partial probability of having shared ingredients with a bias designed to favor such combinations. 
\begin{align}
\beta &= ( \sum_{i=1}^n Shared(x_i,x_{n+1}) )^2 \\
Shared(x_1,x_2) &= \begin{cases}1 & w(x_1,x_2)>0\\0 & else\end{cases}
\end{align}
This process repeated recursively to generate a recipe with the desired number of ingredients.
\begin{align}
p(x_{n+1}) &=\sum_{i=1}^{n}{ \alpha  \beta \frac{w(x_{n+1},v_{j})}{\sum_{j=1}^{k}{w(x_{i}, v_{j})}}}
\end{align}
Finally, resultant combinations are referenced back against the original corpus to guarantee novelty in the result.

\subsection{Neural Language Model}
Our second technique uses a neural language model to generate both the ingredients for a recipe and the steps of the ingredients as well.
We use the same knowledge base as described in Sec. Ingredient Graph~\ref{sec:ingr} and train two separate language models: one to generate the ingredients, and the other to generate the recipe given a set of ingredients.

The first language model uses a simple 4-layer LSTM to generate a sequence of ingredients, treating all the words in a single ingredient as a single token.
For example, ``peanut butter'' would be considered a single token in this model.
We train this model using the sets of ingredients found in each recipe for the entire recipe dataset, with each set ending with an $<$EOI$>$ or End of Ingredients tag.
Once trained, the model then generates a sequence of ingredients until the $<$EOI$>$ is reached using the top-$k$ sampling technique \cite{Holtzman2019TheCC}.

To generate the actual recipe, we use GPT-2 \cite{radford2019language} and fine-tune their pre-trained 345m parameter model on the recipe data.
The data to fine-tune this model is designed to contain the recipe title, ingredients, and instructions in an unstructured text-form.
Once this model has been fine-tuned, we use it to generate the title and instructions for the recipe conditioned on the ingredients generated by the first language model.
The entire generated recipe consists of the ingredients, title, and instructions.

\section{Quest Assembly}

We now use the generated content---i.e. the recipe---to assemble a quest by grounding the generated ingredients and instructions in the game world.
This requires us to first determine the structure of the game world and the locations of objects within this world in addition to transforming the set of generated instructions into executable actions.
We use two types of semantically grounded knowledge graphs to represent this information: the object and action graphs.
\begin{figure}
    \centerline{\includegraphics[width=\linewidth]{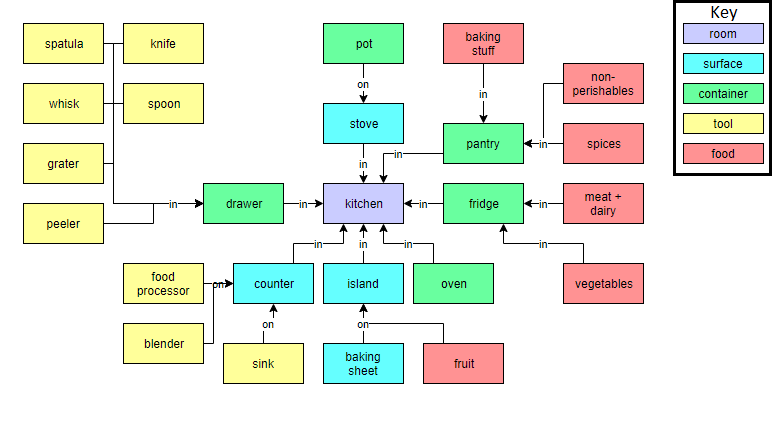}}
    \caption{Object graph in the one room map.}
    \label{1Rmap}
\end{figure}

The object graph is used to determine the structure of the world and the most likely locations of objects within this world.
For example, we could have information that says that vegetables must be stored in a refrigerator.
If a recipe requires carrots, then the carrots would automatically be placed in a refrigerator at the start of the game. 
This graph is constructed by hand and is built to make the game world and resulting quest as coherent as possible.

We construct object graphs for two different room layouts.
The first, the one room (1R) map, consists of a kitchen as well as the objects and actions that exist within it. 
The second map, the five room (5R) map, is an extension of the first map and contains four additional rooms. 

\begin{figure}
    \centerline{\includegraphics[width=\linewidth]{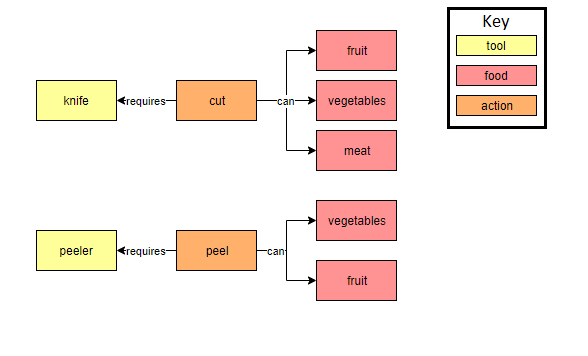}}
    \caption{Example action graph for both maps.}
    \label{actionmap}
\end{figure}

The object graph for the 1R map as shown in Fig.~\ref{1Rmap} is largely inspired by the simple, pre-built game provided within TextWorld \cite{cote2018textworld}. 
This object graph determines how and where objects are placed within the environment during game generation, and the action graph (Fig.\ref{actionmap}) dictates how generated instructions are transformed into executable actions in the game.
The object graph was constructed logically: tools and utensils go in the drawer, meat and dairy belong in the refrigerator, and so on. 
Food item placements are deterministic and coherent. 
Vegetables always go in the refrigerator, and fruit always goes on the kitchen island. 
The action graph was also designed to prevent the player from conducting illogical actions. 

\begin{figure}
    \centerline{\includegraphics[width=0.8\linewidth]{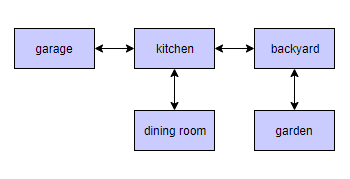}}
    \caption{Room layout in the five room map.}
    \label{roommap}
\end{figure}

\begin{figure}
    \centerline{\includegraphics[width=\linewidth]{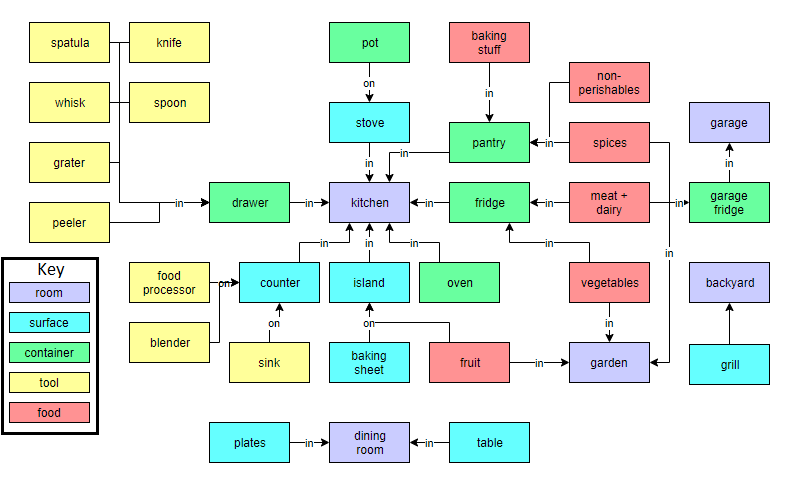}}
    \caption{Object graph in the five room map.}
    \label{5Rmap}
\end{figure}
The 5R map included a dining room, garage, backyard, and garden in addition to the kitchen (Fig.~\ref{roommap}). 
The map (Fig.~\ref{5Rmap}) is designed to maintain the same levels of coherency as the 1R map while allowing for more diverse gameplay, which could in turn lead to higher levels of perceived creativity. 
The additional rooms are selected based on their possible relationships to the domain of food and cooking, and each new room has its own unique objects that players can interact with. 
For example, the garage has an old refrigerator that can be used to store meat.
These new rooms and objects also allow for dynamic food placement. 
Meat can be placed in one of two refrigerators, and fruits and vegetables can possibly be found in the garden. 
The existence of these new locations is not immediately clear to the player.
The garage and backyard are additionally obscured by closed doors, adding to quest complexity. 
While the additional rooms and dynamic food placement allow for more diverse gameplay, they do not sacrifice coherency. 

The action graph contains information regarding the affordances of the objects in the world and what kinds of objects are required to complete a given generated instruction.
For example, if a generated instruction tells us to prepare vegetables, i.e. cut them, then this graph tells us that there must be a knife somewhere in this world.
This graph is partially extracted from static cooking guides online using a mixture of OpenIE~\cite{Angeli2015} and hand-authored rules to account for the irregularities of cooking guides.
An example of an action graph is given in Fig.~\ref{actionmap}.
A player can peel fruit and vegetables, for example, but cannot peel a steak. 
There are also strict rules on what tools are required for certain actions. 
A player can only cut something if they have a knife and can only peel something with a peeler. 
While this restricts how players can interact with the environment, it ultimately reinforces game coherency.

We also note that when generating the quests, both the Markov chain and the neural language model based generation systems use the object graph to determine object placement but only the Markov chain based model uses the action graph.
This is because the instructions generated by the Markov chain model is in the form of a sequence of ingredients which then requires the action graph to determine the actions and additional objects required to turn this list of ingredients into a playable quest.
The action graph would thus take an ingredient such as a carrot and determine first that it needs to be cut and that a knife is required for this task.
The neural language model on the other hand already generates the full action, including potentially required objects, that can be executed and so does not make use of this graph.

\section{Experiments}

Our experiments were designed to compare perceived creativity and coherence, specifically testing our models in addition to factors such as complexity. 
We tested five types of designs: Human Designed~(HD), Random Assignment~(RA), Markov Chains Simple~(MCS), Markov Chains Complex~(MCC), and Language Model~(LM). 
HD is simply what it sounds like: a game that was created by a person.
In this game, a human---not associated with the research---creates both the ingredients and the instructions for a recipe and is additionally responsible for quest assembly, i.e. grounding the generated content in a given game world.
We do not consider experience in designing text-adventure games when picking a human to create this game as this task can be performed even by novices given the easily understandable ``home'' theme of the game world.
The game is manually crafted in terms of decided what ingredients to put where and what the final recipe would come together to be. 
RA is on the opposite end of the spectrum where, as the name suggests, everything is placed in a random location, and the recipe could be totally random with ingredients and instructions that might not normally be seen. 
MCS and MCC use our Markov chains approach to generate quest content.
The difference between MCS and MCC are that the former has four ingredients involved in its recipe while the latter has eight.
This was to vary the complexity to see how that affected perceived creativity.
LM refers to the games generated using the recipes generated by the language model.
We additionally had one-room and five-room variants for each of the models to test how the structure and length of the game would affect the players.

Automatically evaluating the creativity of the output of any computational generation process is a difficult task which requires concrete definitions of the metrics being used.
We thus evaluate by deploying the game designs on Amazon Mechanical Turk for people to play and provide feedback. 
Specifically, they would play one randomly selected game from the 1 room layout and then fill out a survey for that game, and then play one randomly selected game from the 5 room layout and fill out an identical survey.
Subjects were provided with a simple practice game that they could play beforehand to familiarize themselves with TextWorld and its interface.
We had 75 total participants for the entire study and had an average of 15 people play each game.
The only restrictions that we had for participants was that they had to be fluent in English---this was determined by means of pre-built restrictions on Amazon Mechanical Turk and game completion verification.

The users were asked questions pertaining to two metrics: coherence and creativity.
We looked at creativity as a metric in the survey using the components of creativity as defined by Boden: novelty, surprise, and value.
The survey detailed questions that measured our defined metrics, using Likert Scale values along a scale of 1-7. 
It posed questions such as ``How coherent was the objective of the quest?'', ``How original was the quest you played? 1: not at all novel, 7: exceptionally novel'', ``Did you have a sense of accomplishment after completing the game? 1: no value, 7: extremely valuable'', ``How unpredictable was the quest you played?'' when measuring coherence, novelty, value, and surprise respectively.
The other factors were also measured using similarly phrased questions.
A one-way ANOVA test was then conducted followed by Tukey HSD post-hoc analysis to determine significance.
The results of the raw scores for each group as well as the significant results between pairs of different models are presented below.

\section{Results and Discussion}
\label{sec:res}


\begin{figure}[t]
    \centerline{\includegraphics[width=\linewidth]{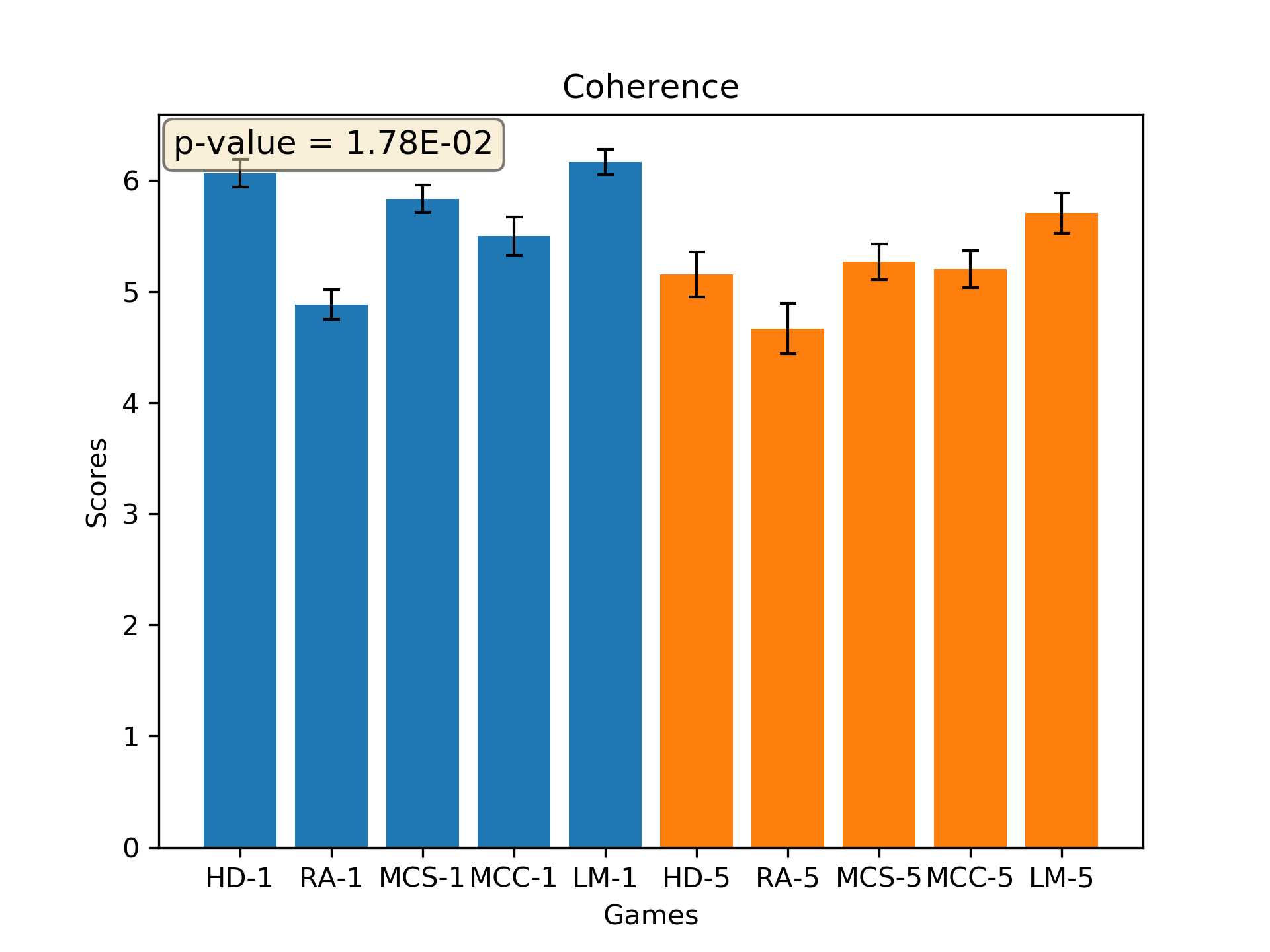}}
    \caption{Coherence scores for each game. Error bars indicate a 95\% confidence interval.}
    \label{fig:coherence}
\end{figure}
\begin{figure}[t]
    \centering
    \includegraphics[width=\linewidth]{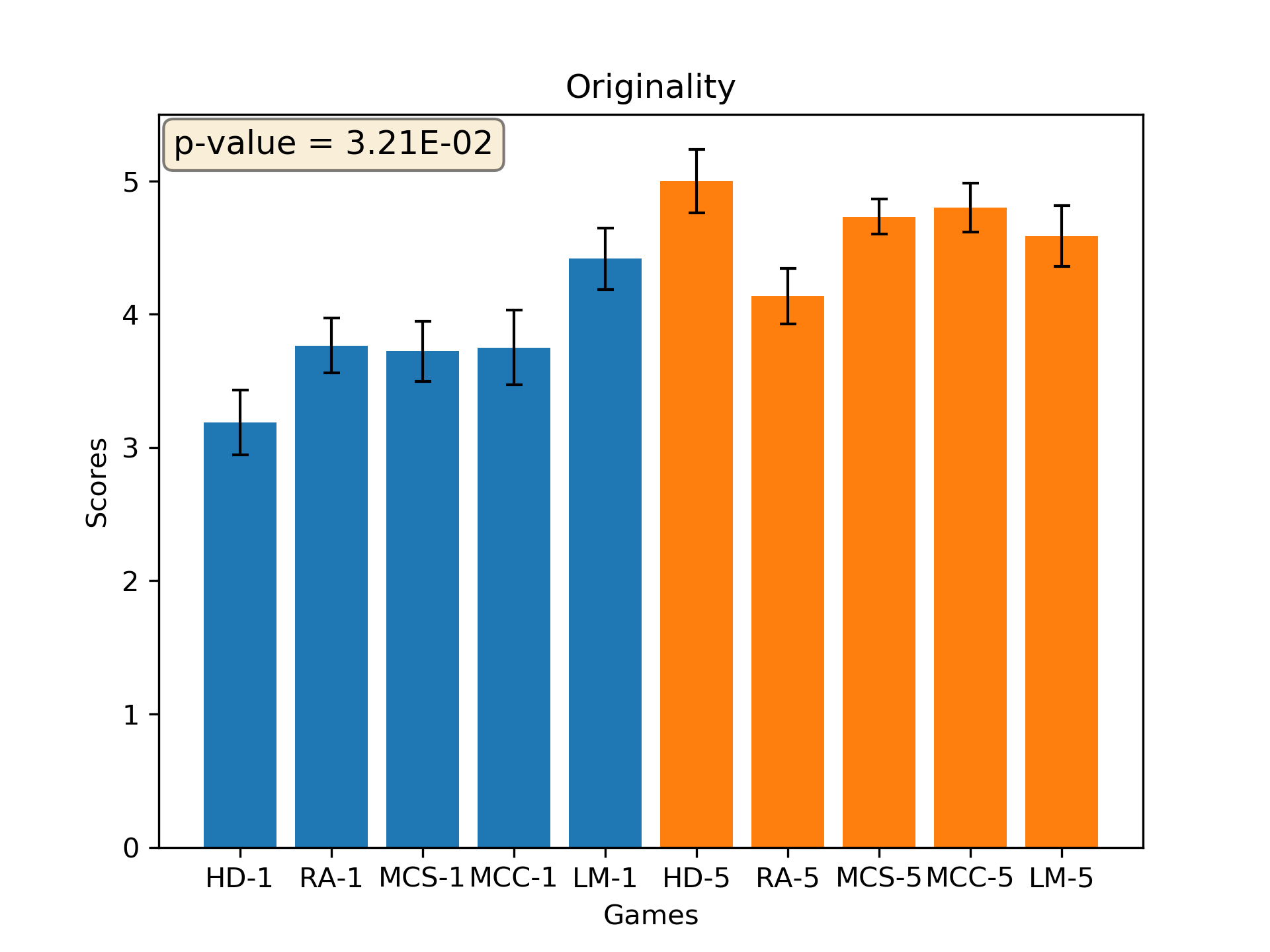}
    \caption{Originality (novelty) scores for each game. Error bars indicate a 95\% confidence interval.}
    \label{fig:originality}
\end{figure}
\begin{figure}[t]
    \centering
    \centerline{\includegraphics[width=\linewidth]{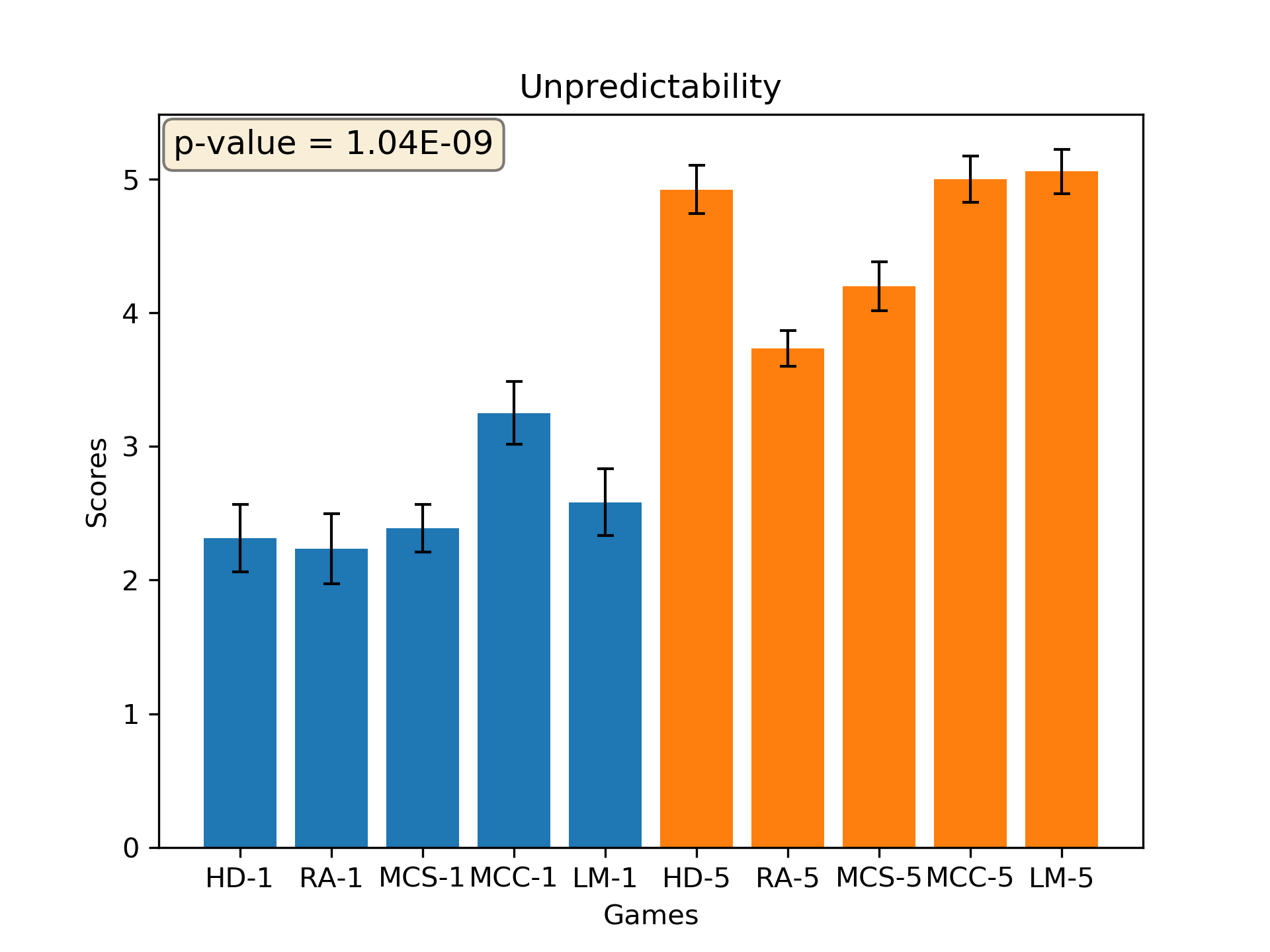}}
    \caption{Unpredictability (surprise) scores for each game. Error bars indicate a 95\% confidence interval.}
    \label{fig:unpredictability}
\end{figure}
\begin{figure}[t]
    \centering
    \centerline{\includegraphics[width=\linewidth]{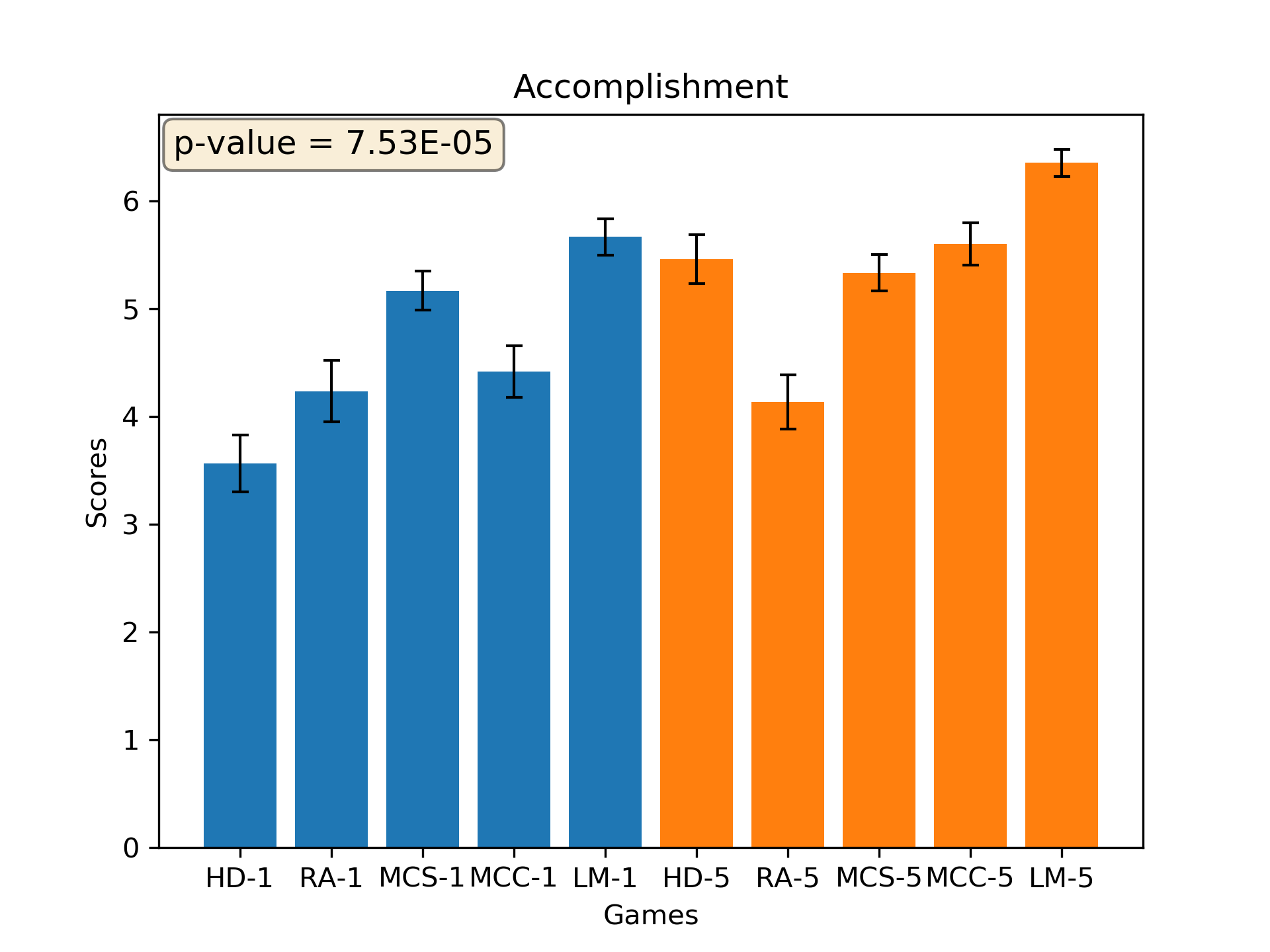}}
    \caption{Accomplishment (value) scores for each game. Error bars indicate a 95\% confidence interval.}
    \label{fig:accomplishment}
\end{figure}

\begin{table}[t]
    \centering
    \begin{tabular}{llrrr}
     \textbf{Group 1} & \textbf{Group 2} & \textbf{Meandiff}
     & \textbf{$p$-value}\\ 
    \hline
     \multicolumn{3}{l}{Coherence}\\
    \hline
MCC-5                                & RA-5                                 & -0.5333                               & 0.1\\
LM-5                                 & MCS-5                                & -0.4392                               & 0.066\\
LM-5                                 & RA-5                                 & -1.0392                               & 0.028\\
LM-1                                 & RA-1                                 & -0.2843      & 0.042 \\                        \hline
    \multicolumn{3}{l}{Originality (novelty)}\\
    \hline
HD-1                                 & LM-1                                 & 1.2292    & 0.057                            \\
HD-1                                 & MCS-1                                & 0.5347     & 0.10                           \\
HD-1                                 & RA-1                                 & 1.5772     & 0.013                           \\
MCC-1                                & RA-1                                 & 1.0147     & 0.077\\                             \hline
   
    \multicolumn{3}{l}{Unpredictability (surprise)}\\
    \hline
        HD-1                                 & MCC-1                                & 0.9375                                & 0.079\\
        HD-1                                 & RA-1                                 & 0.9228                                & 0.071\\
        LM-1                                 & MCC-1                                & 0.6667                                & 0.1\\
        LM-5                                 & RA-5                                 & -1.3255                              & 0.026 \\
        MCC-1                                & MCS-1                                & -0.8611                           &    0.086\\
        MCC-5                                & RA-5                                 & -1.2667                           & 0.037    \\
    \hline
    \multicolumn{3}{l}{Accomplishment (value)}\\
    \hline
HD-1                                 & LM-1                                 & 2.1042                                & 0.022\\
HD-1                                 & RA-1                                 & 1.6728                                & 0.01\\
HD-5                                 & LM-5                                 & 0.8914                                & 0.046\\
LM-1                                 & MCC-1                                & -1.25                                & 0.062 \\
LM-1                                 & MCS-1                                & -0.5                                  & 0.1\\
LM-5                                 & MCS-5                                & -1.0196                               & 0.06\\
MCS-5                                & RA-5                                 & -1.2   & 0.05\\
    \hline
    \end{tabular}
    \caption{Select pairwise results from the post hoc Tukey HSD test for each experiment.}
    \label{tab:results}
\end{table}

We present results for four metrics: coherence, unpredictability (or surprise), novelty (or originality), and value (or accomplishment) for each of the games.
Additionally, we also show the $p$-value result of a one way ANOVA test for the distributions in each of the categories to determine statistical significance.
This test tells us if the differences in the means across the different games are significant for each of the categories separately.
The Tukey HSD post-hoc analysis further tells us which specific pairs of results are significant.
We hypothesized that semantic grounding using the knowledge graph would enable our models to maintain coherence on par with the human designed games.
Further, given the stochastic nature of our generative models, we further predicted that our models would also rate as being comparable in terms of creativity to the human designed games---with all models relatively outperforming the randomly generated games.
We see below that these predictions hold.

We find that the results for each individual category are significant at $p<0.05$ in all the cases.
Additionally, all the specific pairwise comparisons we make are significant with $p<0.1$.
Table~\ref{tab:results} presents some pairwise results along with the corresponding difference in scores and $p$-values.
The rest of this section discusses these metrics in more detail.

Fig.~\ref{fig:coherence} displays trends in the players' perception of coherence for each of the games.
We first see that the one-room games were consistently rated to be more coherent than the five-room games, indicating that overall quest coherence---and thus the coherence of our generative system---degrades the longer and more complex the quest.
Across the games, we see that the RA models were the considered to be the least coherent.
The LM achieves a higher score than both of the Markov chain models and maintains coherence more easily than either.
Most importantly, all of these methods are comparable in coherence to the human-authored games, i.e. our semantically grounded knowledge graph ensures that coherence is not lost when generating content.

Originality (Fig.~\ref{fig:originality}), which we use as a proxy to measure novelty, exhibits similar trends as surprise.
The more longer, more complex games are deemed more original.
Despite being random, the RA games are seen to be less original than the the rest of the games perhaps indicating that there is a link between perceptions of coherence and originality.
The gaps in performance here are much less pronounced, however, and the Markov chain models slightly edge out the LM --- with all three being comparable to the HD games.

Similarly, Fig.~\ref{fig:unpredictability}
describes how surprising the game was to the players.
The difference between the one-room and five-room games here is much more pronounced.
The players find the five-room, the longer and more complex game, much more surprising than their one-room counterparts, showing that complexity is an important factor in determining surprise.
The LM achieves comparable performance to MCC and again they all perform as well as the HD games.

To measure value, or utility, in a text-adventure game, we asked the players if they felt a sense of accomplishment after finishing the game (Fig.~\ref{fig:accomplishment}).
We see players reported a higher sense of accomplishment after finishing more complex games in general with the exception of the RA games, both of which performed poorly---likely due to them being relatively incoherent.
We also note that the LM showed the highest values here, surpassing the HD games.
We hypothesize that this might be due to the player having to perform a wider range of actions, some relatively unintuitive, that are not constrained by our action graph.

\section{Conclusions}

We have demonstrated a framework to automatically generate cooking quests in a ``home'' themed text-adventure game, although our framework can be generalized to other themes as well.
Quest generation in a given game world is a subset of the overall problem of generating entire text-adventure games.
Content generated by both the Markov chains and the neural language models can be grounded into a given game world using domain knowledge encoded in the form of a knowledge graph.
The models each excel on different metrics: the Markov chains model produces quests that are more surprising and novel while the neural language model offers greater value and coherence.
We also note, however, that the neural language model requires less domain knowledge than the Markov chains and is thus potentially more generalizable to other themes and types of quests.

Our human subject study shows us that there is an inverse relationship between creativity and coherence but only when a certain threshold of coherence is passed.
In other words, the less coherent a game the more creative it is, but incoherent games---such as those generated by the RA model---are perceived to be less creative.
Furthermore, our automatically generated games consistently perform at least as well as human designed games in this setting, both in terms of coherence and creativity---implying that the generative process can be automated without a loss in perceived game quality.


\bibliographystyle{iccc}
\bibliography{iccc}


\end{document}